\documentclass[10pt]{article}

\usepackage{float}
\usepackage{rotating}
\usepackage{changepage}

\usepackage[font=normalsize,format=plain,labelfont=bf,up,textfont=normal,up,justification=justified,singlelinecheck=false]{caption}

\usepackage{amssymb}
\usepackage{latexsym}
\usepackage{amsfonts}
\usepackage{amsmath}

\usepackage{theorem}
\newcommand{\BlackBox}{ule{1.5ex}{1.5ex}}  

\usepackage[tight,normalsize,sf,SF]{subfigure}

\usepackage{algorithm}
\usepackage{algorithmic}

\usepackage{xcolor}
\usepackage{stmaryrd}

\newcommand{\eq}[1]{(\ref{#1})}

\newcommand{\inner}[2]{\left\langle #1,#2 \right\rangle}

\newcommand{\Xcal}{\mathcal{X}}
\newcommand{\Ycal}{\mathcal{Y}}

\newcommand{\Lcal}{\mathcal{L}}

\newcommand{\Dcal}{\mathcal{D}}

\newcommand{\x}{\mathbf{x}}
\newcommand{\y}{\mathbf{y}}
\newcommand{\w}{\mathbf{w}}

\newcommand{\f}{\mathbf{f}}

\usepackage{xspace} 
\makeatletter
\DeclareRobustCommand\onedot{\futurelet\@let@token\@onedot}
\def\@onedot{\ifx\@let@token.\else.\null\fi\xspace}

\def\eg{\emph{e.g}\onedot} 
\def\ie{\emph{i.e}\onedot} 
\def\cf{\emph{cf}\onedot}

\def\eg{{e.g}\onedot} 
\def\ie{{i.e}\onedot} 
\def\cf{{cf}\onedot}

\usepackage{multirow}

\usepackage{verbatim}
\usepackage{url}
\usepackage{cancel}
\begin{document}
\title{Bayesian Structured Prediction using\\
Gaussian Processes}

\author{S\'{e}bastien~Brati\`{e}res,
	Novi~Quadrianto
        and~Zoubin~Ghahramani\\
Department of Engineering, University of Cambridge, UK}

\date{}

\maketitle

\begin{abstract}
We introduce a conceptually novel structured prediction model, \emph{GPstruct}, which is kernelized, non-parametric and Bayesian, by design.
We motivate the model with respect to existing approaches, among others, conditional random fields (CRFs), maximum margin Markov networks (M$^3$N), and structured support vector machines (SVMstruct), which embody only a subset of its properties.
We present an inference procedure based on Markov Chain Monte Carlo.
The framework can be instantiated for a wide range of structured objects such as linear chains, trees, grids, and other general graphs. 
As a proof of concept, the model is benchmarked on several natural language processing tasks and a video gesture segmentation task involving a linear chain structure.
We show prediction accuracies for GPstruct which are comparable to or exceeding those of CRFs and SVMstruct.
\end{abstract}


\section{Introduction}
\label{sec:introduction}
Much interesting data does not reduce to points, scalars or single categories. Images, DNA sequences and text, for instance, are not just structured objects comprising simpler indendent atoms (pixels, DNA bases and words). The interdependencies among the atoms are rich and define many of the attributes relevant to practical use. 
Suppose that we want to label each pixel in an image as to whether it belongs to background or foreground (image segmentation), or we want to decide whether DNA bases are coding or not.
The output interdependencies suggest that we will perform better in these tasks by considering the structured nature of the output, rather than solving a collection of independent classification problems.

\begin{figure}[h!]
\center{\includegraphics[scale=0.5]{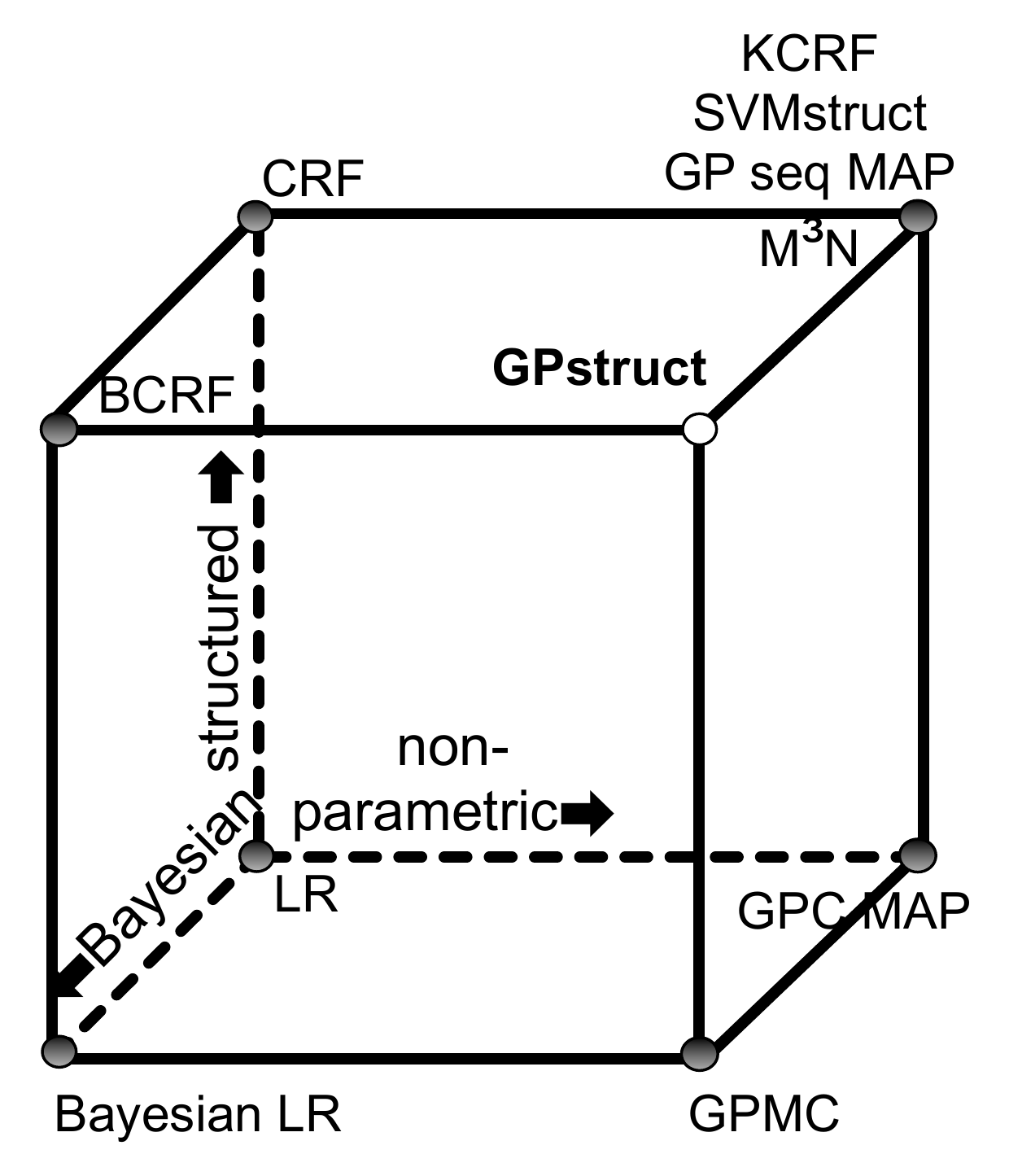}}
  \caption{Schematic representation of structured prediction models. The model we are describing here, GPstruct, exhibits all three properties separately present in other, existing, models.\label{fig:Cube-representation-of}}%
\end{figure}

Existing statistical machine learning models for structured prediction, such as maximum margin Markov Network (M$^3$N) \cite{TasGueKol04}, structured support vector machines (SVMstruct) \cite{TsoJoaHofAlt05} and conditional random field (CRF) \cite{LafMcCPer01}, have established themselves as the state-of-the-art solutions for structured problems (cf. figure \ref{fig:Cube-representation-of} and table \ref{tab:acronyms} for a schematic representation of model relationships). 

\begin{table}[h]
\begin{tabular}{|l|l|l|}
\hline
MRF	& 	Markov random field &  \\
GP	& 	Gaussian process	& \cite{Rasmussen2006}\\
GPMC & 	GP multi-class classification & \cite{Williams1998} \\
CRF & 	conditional random field	& \cite{LafMcCPer01}  \\
KCRF & 	kernelized CRF & \cite{LafZhuLiu04} \\
BCRF & 	Bayesian CRF & \cite{QiSzuMin05}  \\
M$^3$N & 	maximum-margin Markov network & \cite{TasGueKol04} \\
GPC MAP	& GPC maximum a posteriori	&\\
GP seq MAP & GP for sequence labelling & \cite{AltHofSmo04} \\
LR	& 	logistic regression (classification) & \\
SVM	& 	support vector machine &  \\
SVMMC & 	multiclass SVM	& \cite{Weston1999}, \cite{Crammer2001} \\
SVMstruct & 	structured SVM & \cite{TsoJoaHofAlt05} \\
GPstruct & 	structured GP classification & this paper \\
\hline
\end{tabular}
\caption{Models, acronyms and references. A unified view of structured prediction models related to CRF and SVM is given in \cite{PerPonGha07}.  \cite{SutMcC12} presents techniques and applications resulting from a decade of work on CRFs.\label{tab:acronyms}}
\end{table}

In this paper, we focus our attention on CRF-like models due to their probabilistic nature, which allows us to incorporate prior knowledge in a seamless manner. Further, probabilistic models make it possible to compute posterior label probabilities that encode our uncertainty in the predictions.

On the other hand, SVMstruct and M$^3$N offer the ability to use kernel functions for learning using implicit and possibly infinite dimensional features, thus overcoming the drawbacks of finite dimensional parametric models such as the CRF. In addition, kernel combination allows the integration of multiple sources of information in a principled manner. These reasons motivate introducing Mercer kernels in CRFs \cite{LafZhuLiu04}, an advantage that we wish to maintain.

From training and inference point of views, most CRF models estimate their parameters point-wise using some form of optimisation. In contrast, \cite{QiSzuMin05} provide a Bayesian treatment of the CRF which approximates the posterior distribution of the model parameters, in order to subsequently average over this distribution at prediction time. This method avoids important issues such as overfitting, or the necessity of cross-validation for model selection.

Reflecting on this rich history of CRF models, we ask a very natural question: 
can we have a CRF model which is able to use implicit features spaces and at the same time estimates a posterior distribution over model parameters? 
The main drive for pursuing this direction is to combine the best of both worlds from Kernelized CRFs and Bayesian CRFs. 
To achieve this, we investigate the use of Gaussian processes (GP) \cite{Rasmussen2006} for modelling structured data where the structure is imposed by a Markov Random Field as in the CRF.

Our contributions are the following:
\begin{itemize}
  \item a conceptually novel model which combines GPs and CRFs, and its coherent and general formalisation;
  \item while the structure in the model is imposed by a Markov Random Field, which is very general, as a proof of concept we investigate a linear chain instantiation;
  \item a Bayesian learning algorithm which is able to address model selection without the need of cross-validation, a drawback of many popular structured prediction models;
\end{itemize}

The present paper is structured as follows. Section \ref{sec:general-model} describes the model itself, its parameterization and its application to sequential data, following up with our proposed learning algorithm in section \ref{sec:learning}. In section \ref{sec:related-models}, we situate our model with respect to existing structured prediction models.  An experimental evaluation against other models suited to the same task is discussed in section \ref{sec:applications}.

\section{The model}
\label{sec:general-model}
The learning problem addressed in the present paper is structured prediction. Assume data consists of observation-label (or input-output) tuples, which we will note $\Dcal = \{(\x^{(1)},\y^{(1)}), \ldots, (\x^{(N)},\y^{(N)}) \}$, where $N$ is the size of the dataset, and $(\x^{(n)}, \y^{(n)}) \in \Xcal\times\Ycal $ is a data point. In the \emph{structured} context, $\y$ is an object such as a sequence, a grid, or a tree, which exhibits structure in the sense that it consists of interdependent categorical atoms.
Sometimes the output $\y$ is referred to as the \emph{macro-label}, while its constituents are termed \emph{micro-labels}. The observation (input) $\x$ may have some structure of its own. Often, the structure of $\y$ then reflects the structure of $\x$, so that parts of the label map to parts of the observation, but this is not required. The goal of the learning problem is to predict labels for new observations.


The model we introduce here, which we call \emph{\mbox{GPstruct}}, in analogy to the structured support vector machine (SVMstruct) \cite{TsoJoaHofAlt05}, can be succinctly described as follows. 
Latent variables (LV) mediate the influence of the input on the output. The distribution of the output labels given the input and the LV is defined per \emph{clique}: in undirected graphical models, a clique is a set of nodes such that every two nodes are connected. Let this distribution be:
\begin{align}
\label{eq:general-likelihood}
p(\y | \x, \f) = \frac{ \exp (\sum_{c}f(c, \x_c, \y_c))}{\sum_{\y' \in \Ycal}  \exp (\sum_{c}f(c, \x_c, \y'_c))} 
\end{align}
where $\y_c$ and $\x_c$ are tuples of nodes belonging to clique $c$, while $f(c, \x_c, \y_c)$ is a LV associated with this particular clique and values for nodes $\x_c$ and $\y_c$. Let $\f$ be the collection of all LV of the form $f(c, \x_c, \y_c)$. We call the distribution \eq{eq:general-likelihood} \emph{structured softmax}, in analogy to the softmax (a.k.a. multinomial logistic) likelihood used in multinomial logistic regression.
The conditional distribution in \eq{eq:general-likelihood} is essentially a CRF over the input-output pairs, where the potential for each clique $c$ is given by a Gibbs distribution, whose energy function is $E(\x,\y) = \sum_c -f(c, \x_c, \y_c)$. 

In the CRF, potentials are log-linear in the parameters, with basis function $\w^T \phi_c(\x_c, \y_c)$, where $\w$ is the weight parameter and $\phi_c$ a feature extraction function. Here instead, rather than choosing parametric clique potentials as in the CRF, the GPstruct model assumes that $f(c, \x_c, \y_c)$ is a non-parametric function of its arguments, and gives this function a GP prior. Note that we substitute not only $\w$, but the entire basis function by a LV. In particular, $f(c, \x_c, \y_c)$ is drawn from a GP with covariance function (\ie Mercer kernel) $k((c, \x_c, \y_c), (c', \x_{c'}, \y_{c'}))$, so that:
\begin{align}
\f \sim \mathcal{GP} (0,k(\cdot, \cdot))
\end{align}

In summary, the GPstruct is a probabilistic model in which the likelihood is given by a structured softmax, with a Markov random field (MRF) modelling output interdependencies; the MRF's LV, one per factor, are given a GP prior. This MRF could take on many shapes: linear for text, grid-shaped to label pixels in computer vision tasks \cite{NowLam11}, or even, to take a less trivial example, hierarchical, in order to model probabilistic context-free grammars in a natural language parsing task using CRFs \cite{TasKleColKoletal04}. 

\begin{figure}[t]
\begin{center}
\includegraphics[width=0.9\columnwidth]{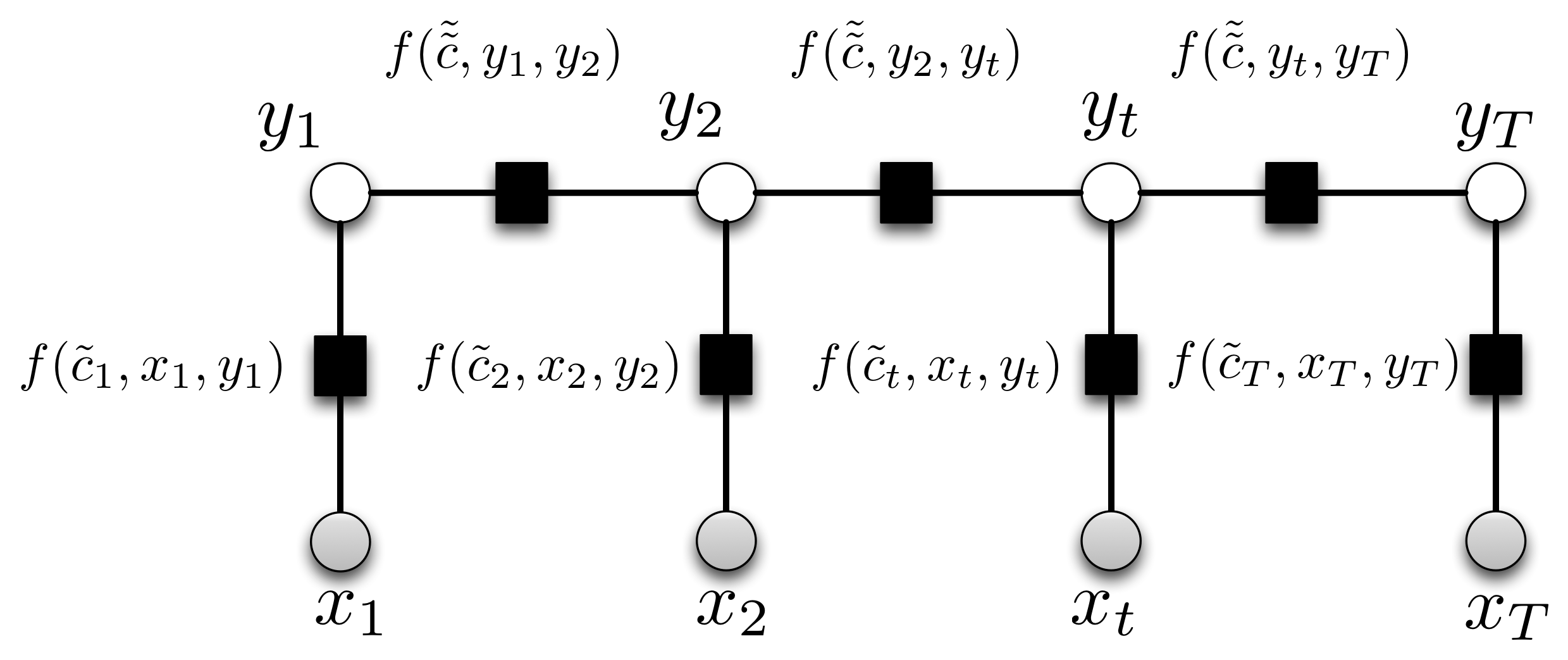}
\end{center}
\caption{Linear-chain factor graph for sequence prediction.}
\label{fig:factorgraph}
\end{figure}

\subsection{Linear chain parameterization}
\label{sub:linear-chain}

While this model is very general, we will now instantiate it for the case of sequential data, on which our experiments are based. Both the input and output consist of a linear chain of equal length $T$, and where the micro-labels all stem from a common set, i.e.  $\Ycal=\underset{t=1, \ldots, T}{\times}\Ycal_t$ with $\forall t, \Ycal_t=\Lcal$ (and the same for $\Xcal$ and $\Xcal_t$). We will therefore write $\y=(y_1, \ldots, y_t, \ldots, y_T)$ and  $\x=(\x_1, \ldots, \x_t, \ldots, \x_T)$. In this context, macro-labels can be called label sequences, and micro-labels just ``labels'', without risk of confusion.

In our experiments below, we tackle text data: the input consists of sentences (chains of words), and outputs of corresponding chains of task-specific micro-labels. A common natural language processing task is word/ sentence/ topic segmentation: here the (2-class) micro-labels are $B$, to label the beginning of a segment, and $I$, to label the inside of a segment. 

We will now expose the design decisions involved in instantiating GPstruct to a linear-chain MRF. A priori, there is one LV per $(c, \x_c, \y_c)$, \ie per clique and node values. 

\emph{Parameter tying} amounts to grouping (tying) some of these LV, thus decreasing the number of LV which need to be learnt. In particular, we will be interested in grouping LV according to their clique type (or \emph{clique template}~\cite{LafZhuLiu04}) or to node values. In the linear chain model, there are two clique types: pairwise cliques $(y_t, y_{t+1})$, and unary cliques $(\x_t, y_t)$. 

In our treatment of linear chain GPstruct, we tie LV as follows: we distinguish each individual unary clique, but tie all pairwise cliques, so that we can denote them by $\tilde{c}_t$ resp. $\tilde{\tilde{c}}$; further, we do not tie based on node values. This is illustrated in figure \ref{fig:factorgraph}. By distinguishing each unary clique, we obtain a non-parametric model, where the number of unary LV grows with the data. Alternatively, we could decide to group all $\tilde{c}_t$ to $\tilde{c}$, or maybe to group all $\tilde{c}_t$ except for the edge positions $t=1$ or $t=T$, where the task may dictate a special behaviour. The same goes for the pairwise LV, where alternative choices may be relevant to the domain. Parameter tying for the linear chain is essentially the same as \eg for a grid.

Let $T^{(n)}$ denote the length of $\y^{(n)}$ (which need not be constant across label sequences). In our chosen parameterization, there is one unary LV for each position $t$, and each value $y_{t}$, so there are $ \sum_n {T^{(n)}} \times \left|\Lcal\right| $ unary LV. This number usually dominates the number of pairwise LV. 

Note that we had to define LV for all possible labels $y_t$ (more generally, $\forall \y \in \Ycal$), not just the ones observed. This is because in \eq{eq:general-likelihood}, the normalisation runs over $\y' \in \Ycal$, and also because we want to evaluate $p(\y | \x, \f)$ for any potential $\y$.
By contrast, the input $\x$ and therefore $x_t$ is always assumed observed in our supervised setting, and so we do not need to define LV for ranges of $x_t$ values.

Now turning to pairwise LV: 
there is one pairwise LV per $(y_{t},y_{t+1})$ tuple; and so there are $ \left|\Ycal_t\right| \times \left|\Ycal_{t+1}\right| = \left|\Lcal\right|^2$ pairwise LV.


\subsection{Kernel function specification}
\label{sec:kernel-spec}
The Gaussian process prior defines a multivariate Gaussian density over any subset of the LV, with usually zero mean and a covariance function given by the positive definite kernel (Mercer kernel) $k$ \cite{SchSmo01}. Our choice of kernel decomposes into a unary and pairwise kernel function: 
\begin{align}
& k((c, \x_c, \y_c),(c', \x_{c'}, \y_{c'})) = \nonumber \\
& \hspace{1cm} \mathbb{I}[c,c' \in \{\tilde{c}_t | t\}] \; k_u((t, \x_t, y_t),(t', \x_{t'}, y_{t'})) \nonumber\\
& \hspace{1cm} + \mathbb{I}[c=c'=\tilde{\tilde{c}}] \; k_p((y_{s},y_{s+1}),(y_{s'},y_{s'+1}))
\end{align}
In the above, we make use of Iverson's bracket notation: $\mathbb{I}[P]=1$ when condition $P$ is true and $0$ otherwise.
The positions of $c,c'$ are noted $t,t'$, and $\x_t,\x_{t'},y_t,y_{t'}$ are the corresponding input resp. output node values. 

We give the unary kernel the form
\begin{align}
\label{eq:kernel-unary}
k_u((t, \x_t, y_t),(t', \x_{t'}, y_{t'})) =
\mathbb{I}[y_{t}=y_{t'}]k_{x}(\x_{t},\x_{t'}).
\end{align}
$k_x$ is an ``input-only'' kernel, for instance the linear kernel $\left\langle \x_{t}, \x_{t'}\right\rangle $, or the squared exponential kernel, defined as the inverse of the exponentiated Euclidean distance: $\exp(-\frac{1}{\gamma}|| \x_{t} - \x_{t'} ||^2)$, where $\gamma$ is a kernel hyperparameter.

Further, our pairwise kernel takes on the form
\begin{align}
\label{eq:kernel-pairwise}
k_{p}((y_{t},y_{t+1}),(y_{t'},y_{t'+1}))= 
\mathbb{I}[y_{t}=y_{t'} \wedge y_{t+1}=y_{t'+1} ]
\end{align}

With the proper ordering of LV, the Gram matrix has a block-diagonal structure:
\begin{align}
K = cov [\f] =
\left(\begin{array}{cc}
K_{\mbox{unary}} & 0\\
0 & K_{\mbox{pairwise}}
\end{array}\right)
\end{align}
It is a square matrix, of length equal to the total number of LV $\sum_n {T^{(n)}} \times \left|\Lcal\right| + |\Lcal|^2$.
$K_{\mbox{unary}}$ is block diagonal, with $|\Lcal|$ equal square blocks, each the Gram matrix of $k_x$ of size $ \sum_n {T^{(n)}}  $,
and $K_{\mbox{pairwise}}=\mathbf{I}_{\left|\Lcal\right|^2}$. 

Summing up sections \ref{sec:general-model} to \ref{sec:kernel-spec}, designing an instance of a GPstruct model requires three types of decisions: the choice of the MRF, mainly dictated by the task and the output data structure; parameter tying; kernel design. The next section now details attractive properties of this model.

%

\subsection{Model properties}
\label{sec:fourprops}
The GPstruct model has the following appealing statistical properties:

\textbf{Structured:} the output structure is controlled by the design of the MRF, which is very general. The only practical limitation is the availability of efficient inference procedures on the MRF.

\textbf{Non-parametric:} the number of LV grows with the size of the data. In the linear chain case, it is the number of unary LV which grows with the total length of input sequences. 

\textbf{Bayesian:}  this is a probabilistic model that supports Bayesian inference, with the usual benefits of Bayesian learning. At prediction time: error bars and reject options. At learning time: model selection and hyperparameter learning with inbuilt Occam's razor effect, without the need for cross-validation. 

\textbf{Kernelized:} a joint (input-output) kernel is defined over the LV. Kernels potentially introduce several hyperparameters, making grid search for cross-validation intractable. Kernelized Bayesian models like GPstruct do not suffer from this, as they define a posterior over the hyperparameters.

\section{Learning}
\label{sec:learning}
Our learning algorithms are Markov chain Monte Carlo procedures, and as such are ``any-time'', in that they have no predefined stopping criterion.

\subsection{Prediction}
\label{sec:prediction}
Given an unseen test point $\x_*$, and assuming the LV $\f_*$ corresponding to $x_*$ to be accessible, we wish to predict label $\hat{\y}_*$ with lowest loss. Now, given $\f_*$, the underlying MRF is fully specified. In tree-shaped structures, belief propagation gives an exact answer in linear time $O(T)$; in the linear chain case, under 0/1 loss $\ell(\y,\hat{\y})=\delta(\y=\hat{\y})$, we predict the jointly most probable output sequence obtained from the Viterbi procedure, and under Hamming (micro-label-wise) error $\ell(\y,\hat{\y})=\sum_t \delta(\y_t=\hat{\y}_t)$, we predict the micro-label-wise most probable output sequence. 
For other cases than trees, where exact inference is impossible, approximate inference methods such as loopy belief propagation \cite{murphy1999loopy} are available. 

Given $\f$, due to the GP marginalisation property, the test point LV $\f_*$ are distributed according to a multivariate Gaussian distribution (\cf\ \eg \cite[section 2]{Nickisch2008} for a derivation): 
$\f_* | \f \sim N(K_*^T K^{-1} \f,K_{**} - K_*^T K^{-1} K_*)$,
where matrices $K, K_*, K_{**}$ have their element $(i,j)$ equal to $k(\f^{i}, \f^{(j)})$ resp. $k(\f^{i}, \f_*^{(j)})$ resp. $k(\f_*^{(i)}, \f_*^{(j)})$, with $k$ the kernel described section \ref{sec:kernel-spec}.   

Uncertainty over $\f_* | \f$ is accounted for correctly by \emph{Bayesian model averaging}:
$\hat{\y}_* = \arg \max_{\y_* \in \Ycal_*} p(\y_* | \f)$, with 
\vspace{-0.14cm}
\begin{align}
\label{eq:bma-f*}
p(\y_* | \f) = \frac{1}{N_{\f_* | \f}} \sum_{\f_* | \f} p(\y_* | \f_*)
\end{align}
where $N_{\f* | \f}$ is the number of samples from $\f_* | \f$. 

\subsection{Sampling from the posterior}

We wish to represent the posterior distribution $\f | \Dcal$ (as opposed to performing a MAP approximation of the posterior to a single value $\f_{MAP}$). The training data likelihood is $p(\Dcal | \f) = \prod_{n} p(\y^{(n)} | \f, \x^{(n)})$, with the single point likelihood given by \eq{eq:general-likelihood}. Training aims at maximizing the likelihood, for which we propose to use elliptical slice sampling (ESS) \cite{MurAdaMac10}, an efficient MCMC procedure for ML training of tightly coupled LV with a Gaussian prior. In all our experiments below, we discard the first third of the samples before carrying out prediction, to allow for burn-in of the MCMC chain.

The computation of the likelihood itself is a non-trivial problem due to the presence of the normalising constant, which ranges over $\y' \in \Ycal$, of size exponential in the number of micro-labels $|\Lcal|$. In the case of tree-shaped MRFs, however, belief propagation yields an exact and usually efficient procedure; in the linear case, it is referred to as forwards-backwards procedure, and runs in $O(T|\Lcal|^2)$.

ESS requires computing the full kernel matrix, of size $O(N_{LV}^2)$, where $N_{LV}$ is the total number of LV, and its Cholesky, obtained in $O(N_{LV}^2)$ time steps. The large size of the matrices is a limiting factor to our implementation.  

ESS yields samples of the posterior. To perform prediction, it is necessary to introduce one more level of Bayesian model averaging: continuing from \eq{eq:bma-f*}, $p(\y_* | \Dcal) = \frac{1}{N_{\f}} \sum_{\f} p(\y_* | \f)$
where $N_{\f}$ is the number of samples of $\f | \Dcal$ available.

\section{Relation to other models}
\label{sec:related-models}
Our proposed method builds upon a large body of existing models, none of which, however, exhibit all properties mentioned in section \ref{sec:fourprops}. 

\subsection{GP classification}

Gaussian process models (or any regression models such as a linear regression) can be applied to classification problems. 
In a probabilistic approach to classification, the goal is to model posterior probabilities of an input point $\x$ belonging to one of $|\Lcal|$ classes, 
$y\in \{ 1,\ldots, |\Lcal|\}$. 
For binary classification (that is $|\Lcal|=2$), we can turn the output of a Gaussian process (in $\mathbb{R}$) into a class probability (in the interval $[0,1]$) by using an appropriate non-linear activation function.  
The most commonly used such function is the logistic function
$p(y = 1|f,\x)  = \frac{\exp(f(\x))}{\exp(f(\x)) + \exp(-f(\x))}$.
The resulting classification model is called \emph{GP binary classification} \cite{Williams1998}.
Let us now move from binary classification to multi-class classification. 
This is achieved by maintaining $K$ regression models, each model being indexed by a latent function $f_k$. 
We use the vector notation $\f = (f_1 \ldots f_K) $ to index the collection of latent functions. The desired multi-class model is obtained by using a generalisation of the logistic to multiple variables, the softmax function: 
$p(y=k |  \f,\x ) = \frac{\exp(f_k(\x))}{\sum_{k=1}^K \exp(f_k(\x))}$.
The corresponding model is called \emph{Gaussian process multi-class classification} (GPMC) \cite{Williams1998}. Note that the above multiclass distribution is normalised over the set of possible output labels $\Lcal$ (here $|\Ycal| = K$). Simply extending the multi-class model for a structured prediction case is computationally infeasible due to the sheer size of the label set $\Lcal$. We provide a novel extension of Gaussian process for structured problems.
Structured prediction itself has a long history of successful methods, which we discuss in subsequent sections.

\subsection{From logistic to structured logistic}
A natural way to cater for interdependencies between micro-labels at prediction time, is to define an MRF over $\y$, and to condition the resulting distribution on the input $\x$ (i.e. in a graphical model representation, inserting directed edges from input to output): we thus obtain a mixed graphical model, the \emph{conditional random field} (CRF) \cite{LafMcCPer01}, a very popular and successful model for structured prediction. The CRF defines a log-linear model for $p(\y|\x)$:
$
p(\y|\x,\w) = \frac{1}{Z(\x,\w)}\exp\left(\sum_c\inner{\w_c}{\phi(\x_c,\y_c)}\right),
$
for a weight vector $\w$, and a joint input-output feature representation $\phi(\x,\y)$. In the above, $Z(\x,\w) = \sum_{\y'\in\Lcal}\prod_{c}\exp(\inner{\w_c}{\phi(\x_c,\y_c)})$ is the normalising constant.
As before, we use $c$ to denote a clique. Instead of parameterizing the energy function, $E(\x,\y) := -\inner{\w_c}{\phi(\x_c,\y_c)}$, by means of a weight vector, GPstruct place a Gaussian process prior over energy functions, effectively side-stepping parameterization.
Recent advances in CRF modelling by \cite{JacNowShaRot12} also side-step linear parameterization of the energy function. Instead, a random forest is used to model the energy function, allowing higher order interactions. 

\subsection{Kernelizing structured logistic}
\cite{LafZhuLiu04} presents a kernelized variant of the CRF, the \emph{kernel conditional random field} (KCRF), where a kernel is defined over clique templates. 
The kernelized version of the CRF is generally difficult to construct, to train, and have several hyperparameters that need to be set via cross-validation, therefore, have not been adopted as enthusiastically as regular CRF.
\emph{Structured support vector machines} \cite{TsoJoaHofAlt05}, SVMstruct, and \emph{maximum margin Markov networks} \cite{TasGueKol04}, M$^3$N also model $p(\y|\x)$ as a log-linear function. However, to learn $\w$, while traditional CRF learning maximizes the conditional log likelihood of the training data, both SVMstruct and M$^3$N perform maximum margin training: learn $\w$ which predicts the correct outputs with a large margin compared to incorrect outputs (all other outputs except the correct ones). SVMstruct can be easily kernelized by means of the representer theorem. Our proposed GPstruct is also kernelized, with a practical advantage that kernel hyperparameters can be inferred from the data instead of requiring a cross-validation procedure.

\subsection{Bayesian versus MAP inference}
By Bayesian inference (as opposed to maximum likelihood ML or maximum a posteriori MAP inference), we mean the preservation of the uncertainty over LV, that is their representation, not as point-wise estimates, but as random variables and their probability distribution.   

CRF parameters are usually estimated point-wise, \eg often with an ML or MAP objective using gradient ascent or approximate likelihood techniques, \cf \cite{SutMcC12} for a review.
An exception to this is \emph{Bayesian conditional random field} (BCRF) proposed by \cite{QiSzuMin05}. 
Instead of point-wise parameter estimation, the BCRF approximates the posterior distribution of the CRF parameters as Gaussian distributions and learns them using expectation propagation \cite{Minka01}. GPstruct also follows a Bayesian inference procedure, and combines it with kernelization. 

Despite the similarity in name, the model in \cite{AltHofSmo04} is more similar to the KCRF than to GPstruct. Like KCRF, this work tackles sequence labelling, while we purposefully formulate GPstruct to apply to any underlying MRF, even though we demonstrate its instantiation in sequences.
More importantly, \cite{AltHofSmo04} take a MAP estimation of the LV, making the model, among others, unable to infer associated hyperparameters directly from the data. However, the point-wise MAP estimate comes with a benefit: sparsification, due to the applicability of the representer theorem. The LV appear as a weighted sum of kernel evaluations over the data. Two methods  are applicable from here. The first, applied in \cite[4.2]{AltHofSmo04} and \cite[3]{LafZhuLiu04}, consists in greedily selecting the LV/ clique associated to the direction of steepest gradient, during the optimisation phase. The second method consists of applying the ``Taskar trick" \cite{TasGueKol04}, and is concerned with the fact that in the LV expression obtained from the representer theorem, the sum runs over $\Ycal$, i.e. all possible macro-labels, which is exponentially large in $|\Lcal|$. This trick consists in a rearrangement of terms inside the objective functions which allows a lower-dimensional reparameterization. These sparsification techniques are not accessible to us due to the use of a Bayesian representation; however alternative techniques may come from the GP sparsification literature.

\subsection{Structured prediction via output kernels}
All previously mentioned structured prediction methods explicitly model the output interdependency via MRF. 
A different strand of work aims at building an implicit model of output correlations via a kernel similarity measure \cite{WesChaEliSchetal02, BoSmi10}.
The twin Gaussian processes of \cite{BoSmi10} address structured continuous-output problems by forcing input kernels to be similar to output kernels.
This objective reflects the assumption that similar inputs should produce similar outputs.
The input and output are separately modelled by GPs with different kernels. Learning consists of minimising KL divergence.

\section{Applications}
\label{sec:applications}
In order to appreciate how the proposed model and learning scheme compare to existing techniques, we conducted benchmark experiments on a range of language processing tasks: segmentation, chunking, and named entity recognition, as well as on a video processing task, gesture segmentation, all involving a linear chain structure.

\subsection{Text Processing Task}
Our data and tasks comes from the CRF++ toolbox\footnote{by Taku Kudo \url{http://crfpp.googlecode.com/svn/trunk/doc/index.html}}. Four standard natural language processing tasks are available, \cf table \ref{tab:nlp}. 
Noun phrase identification (\texttt{Base NP}) tags words occurring in noun phrases with $B$ for beginning, $I$ for a word inside a noun phrase, and $O$ for other words. 
\texttt{Chunking} (\ie shallow parsing) labels sentence constituents. The \texttt{Segmentation} task identifies words (the segments) in sequences of Chinese ideograms. Japanese named entity recognition (\texttt{Japanese NE}) labels several types of named entities (organisation, person, etc.) occurring in text.

The data was used in pre-processed form, with sparse binary features extracted for each word position in each sentence. Results were averaged over five experiments per task. Each experiment's training and test data was extracted from the full data set (sizes given in table \ref{tab:nlp}) so that the training sets were always disjoint -- except in the case of segmentation, a small-data set of 36 sentences overall, which was subjected to five random splits. 

\textbf{Baselines}
We compared GPstruct to CRF and SVMstruct. The CRF implementation\footnote{by Mark Schmidt \url{http://www.di.ens.fr/~mschmidt/Software/crfChain.html}} used LBFGS optimisation. In nested cross-validation, the $L_2$ regularisation parameter ranged over integer powers from $10^{-8}$ to $1$. Prediction in the CRF and GPstruct minimised Hamming loss (\cf section \ref{sec:prediction}).
The SVMstruct\footnote{by Thorsten Joachims \url{http://www.cs.cornell.edu/people/tj/svm_light/svm_hmm.html}} used a linear kernel, for comparison with the CRF.
The regularisation parameter in nested cross-validation ranged over integer powers from $10^{-3}$ to $10^2$.

\textbf{Computing} The CRF package is MEX-compiled Matlab, while the SVMstruct system is coded in C++. Our Matlab implementation of GPstruct used MEX functions from the UGM toolbox\footnote{also by Mark Schmidt \url{http://www.di.ens.fr/~mschmidt/Software/UGM.html}} for likelihood (implementing the forward-backward algorithm). To illustrate runtimes, a 10 hour job on a single core of a 12-core Hex i7-3930K 3.20 GHz machine can accommodate CRF/ SVMstruct learning and prediction, including nested cross-validation over the parameter grid mentioned above, for one experiment, for one task. In the same computing time, GPstruct can perform 100 000 iterations for one experiment for the chunking or segmentation task (the fastest), including hyperparameter sampling (50 000 resp. 80 000 iterations for \texttt{Base NP} resp. \texttt{Japanese NE}). Getting a precise runtime comparison of CRF, SVMstruct and GPstruct code is not straightforward since implementation languages differ. Having said that, our GPstruct experiments were roughly a factor of two slower than the baselines including grid search.

\textbf{Kernel hyperparameter learning}
The GP prior over $\f$ is parameterized by its mean (zero in our case), and the kernel function, which may possess hyperparameters. To explore the effect of kernel hyperparameter learning, we introduce a multiplicative hyperparameter $h_p$ in front of the pairwise kernel, and give it a scaled Gamma hyperprior : $h_p / 10^{-4} \sim  \mathrm{Gamma}(1,2)$. 

MCMC sampling of the hyperparameter is performed using the prior whitening technique \cite{MurAdaMac10}, which is easy to implement. Surrogate data modelling \cite{MurAdaMac10} is tailored to GP prior LV models, and is reported to give better results; however, it requires an approximation of the posterior variance for the structured softmax case. While it is possible to derive such approximations, we could not observe any performance gain in our experiments so far. 

Experimentally, ESS (which samples $\f | \Dcal$) needs to be run over many more iterations than hyperparameter sampling (sampling from the hyperparameter posterior $\mathbf{h} | \Dcal$). We therefore sample from the hyperparameter once every 1 000 ESS steps. Kernel learning is possible as well with GPstruct, but a few exploratory experiments using polynomial and squared exponential kernels on the binary-valued text datasets did not improve the performance.
 
\begin{table*}[tb]
\begin{adjustwidth}{-1in}{-1in}
\centering
\begin{tabular}{|l|*{5}{r@{$\pm$}l|}}
     & \multicolumn{2}{c|}{{\tt Base NP}} &\multicolumn{2}{c|}{{\tt Chunking}} & \multicolumn{2}{c|}{{\tt Segmentation}}  & \multicolumn{2}{c|}{{\tt Japanese NE}}\\
number of categories          & \multicolumn{2}{c|}{{ 3}} &\multicolumn{2}{c|}{{ 14}} & \multicolumn{2}{c|}{{ 2}}  & \multicolumn{2}{c|}{{ 17}}\\
number of features & \multicolumn{2}{c|}{{ 6,438}} &\multicolumn{2}{c|}{{ 29,764}} & \multicolumn{2}{c|}{{ 1,386}}  & \multicolumn{2}{c|}{{ 102,799}}\\
size training/ test set (sentences) & \multicolumn{2}{c|}{{ 150 / 150}} &\multicolumn{2}{c|}{{ 50 / 50}} & \multicolumn{2}{c|}{{ 20 / 16}}  & \multicolumn{2}{c|}{{ 50 / 50}}\\
     
\hline\hline
{\bf SVMstruct}      			& 5.91&0.43		& 9.79&0.97		& 16.21&2.21		& 5.64&0.82\\
{\bf CRF}      	 		& 5.92&0.23 	& \bf{8.29}&\bf{0.76}	& 14.98&1.11	& \bf{5.11}&\bf{0.65}\\
{\bf GPstruct ($h_p = 1$)}      	& \bf{4.81}&\bf{0.47}     & 8.76&1.08  	& 14.87&1.79 		& 5.82&0.83 \\
{\bf GPstruct (prior whitening)}	& 5.06&0.38   	& 8.57&1.20  	& \bf{14.77}&\bf{1.78} 	& 5.65&0.80 \\

\end{tabular}
 \end{adjustwidth}

\caption{Experimental results on text processing task. Error rate across $5$ experiments (mean $\pm$ one standard deviation). GPstruct experiments on 250 000 ESS steps (\ie $\f$ samples), using the $\f_*$ MAP scheme, linear kernel, sampling hyperparamers every 1 000 steps (prior whitening) or fixing $h_p=1$, thinning at 1:1 000.}
    \label{tab:nlp}
\end{table*}

\textbf{Results and interpretation}
Our experimental results are summarised in table \ref{tab:nlp}. 
GPstruct is generally comparable to the CRF, and slightly better than SVMstruct.
Our choice of hyperprior does not seem to fit the Base NP task, where hyperparameter sampling turns out to be worse than keeping $h_p$ fixed at 1.

\subsection{Video Processing Task}

\begin{figure}[tb]
\centering
  \includegraphics[scale=0.6]{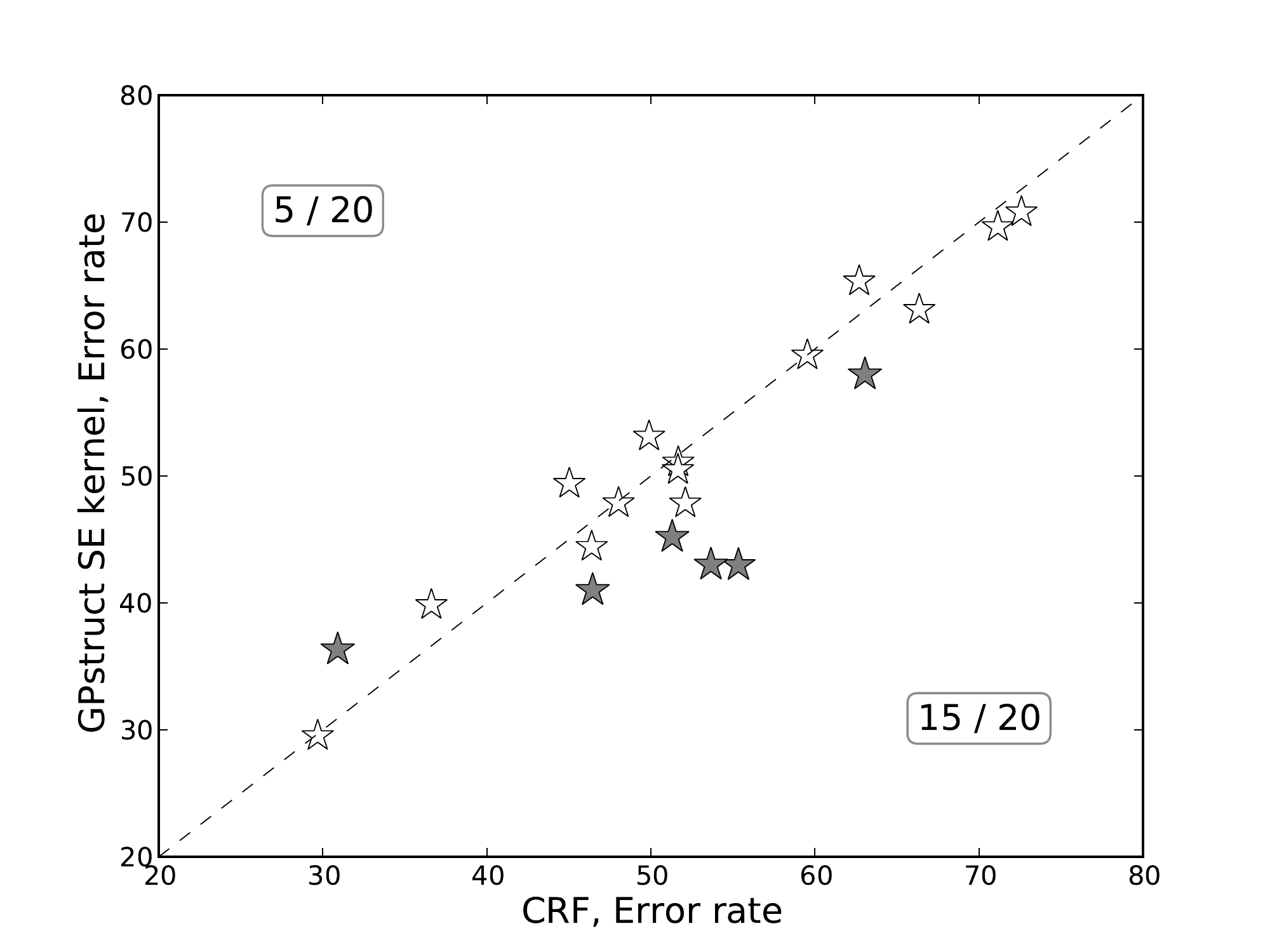}
  \caption{Error rate cross plot of the $20$ gesture video sessions. The axes correspond to error rate of GPstruct with SE kernel and CRF, the diagonal line shows equal performance. The shadowed stars are those with at least $5\%$ performance difference. \label{fig:performance}}%
\end{figure}

In a second set of experiments, we apply CRF and GPstruct to the ChaLearn gesture recognition dataset\footnote{\url{https://sites.google.com/a/chalearn.org/gesturechallenge/}}. The data in this case consists of video sequences of an actor performing certain gestures. Each video frame is labelled with a gesture. The videos have an average length of 86 frames, and maximum length 305 frames. The dataset has 20 sessions of 47 videos each. The label space size varies for different sessions, between 9 and 13. For each session, we use 10 videos to train a chain CRF or GPstruct and the rest as test data. At each frame of the video we extract HOG/HOF \cite{LapMarSchRoz08} descriptors and construct a codebook of 30 visual words using a $k$-means clustering algorithm. Frames are represented by normalized histograms of 
visual words occurrence, resulting in $30$ feature dimensions.
A squared exponential kernel, $\exp(-\frac{1}{\gamma}|| \x_{t} - \x_{t'} ||^2)$, was used. The kernel hyperparameter 
$\gamma$ is given a $\mathrm{Gamma}(1,2)$ prior and is initially set to the median pairwise distance.

\textbf{Results}
The experimental results summarize as follows: averaged across all 20 sessions, the error rates were $52.12 \pm 11.73$ for the CRF, $51.91\pm11.02$ for GPstruct linear kernel, and $50.42 \pm 11.24$ for GPstruct SE kernel. Since each session effectively represents one specific learning task, we compare the pairwise performances across $20$ sessions between GPstruct SE kernel and CRF in figure \ref{fig:performance}. GPstruct outperforms the CRF baseline by more than $5\%$ in five cases, while it underperforms it in one case. The performance between GPstruct linear kernel and CRF are comparable and we did not include details here due to space constraints.

\subsection{Practical insights}

We will open-source our GPstruct code on MLOSS\footnote{\url{www.mloss.org}} to expose the GPstruct model more widely and encourage experimentation.

\textbf{Kernel matrix positive-definiteness} To preserve numerical stability of the Cholesky operation, diagonal jitter of $10^{-4}$ is added to the kernel matrices. Depending on the hyperprior, some hyperparameter samples may make the kernel matrices badly conditioned: this is best prevented by rejecting such a proposal by simulating a very low likelihood value.

\begin{figure*}[p]
\centering
  \includegraphics[width=0.65\columnwidth]{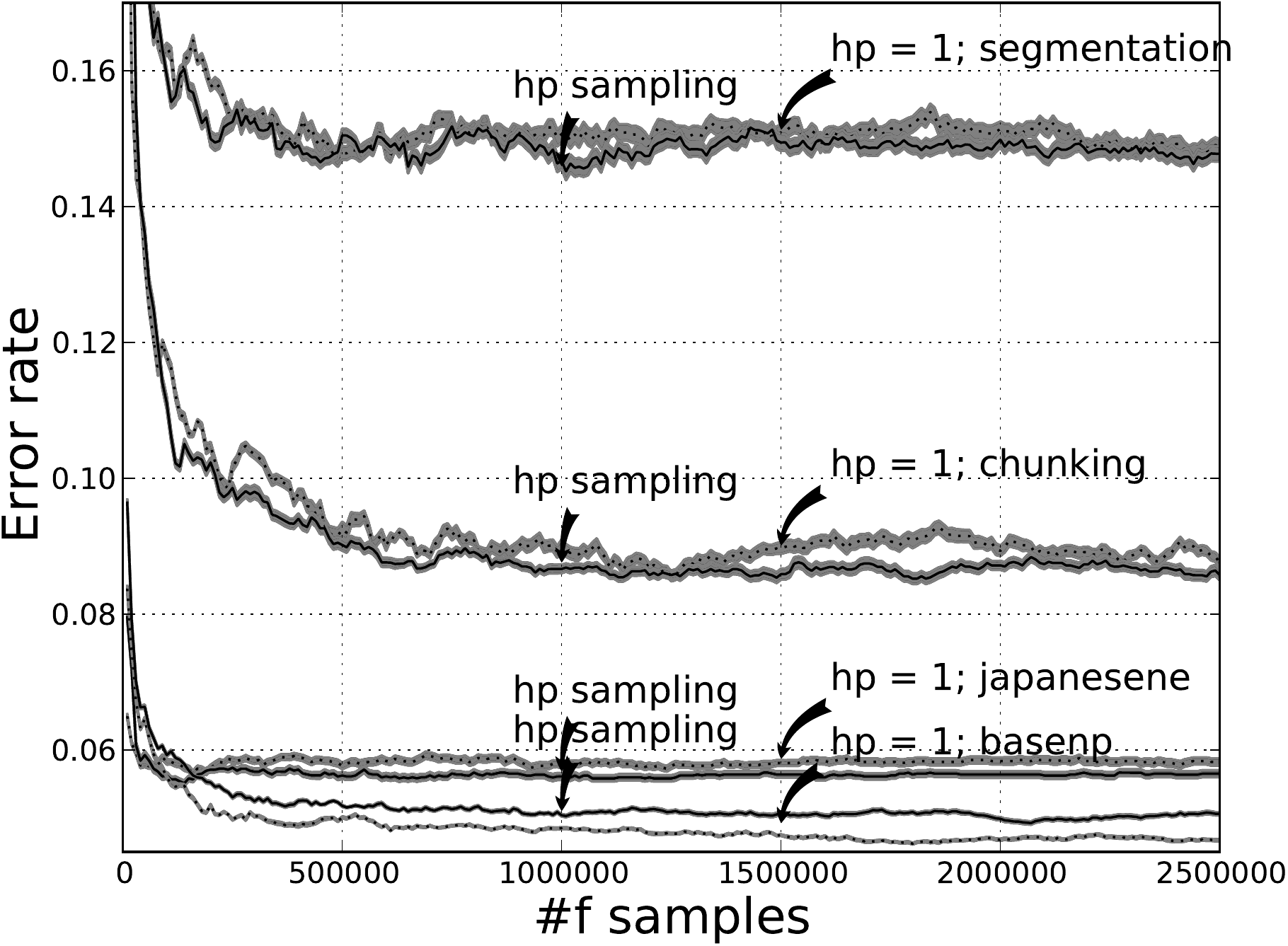}
  \includegraphics[width=0.65\columnwidth]{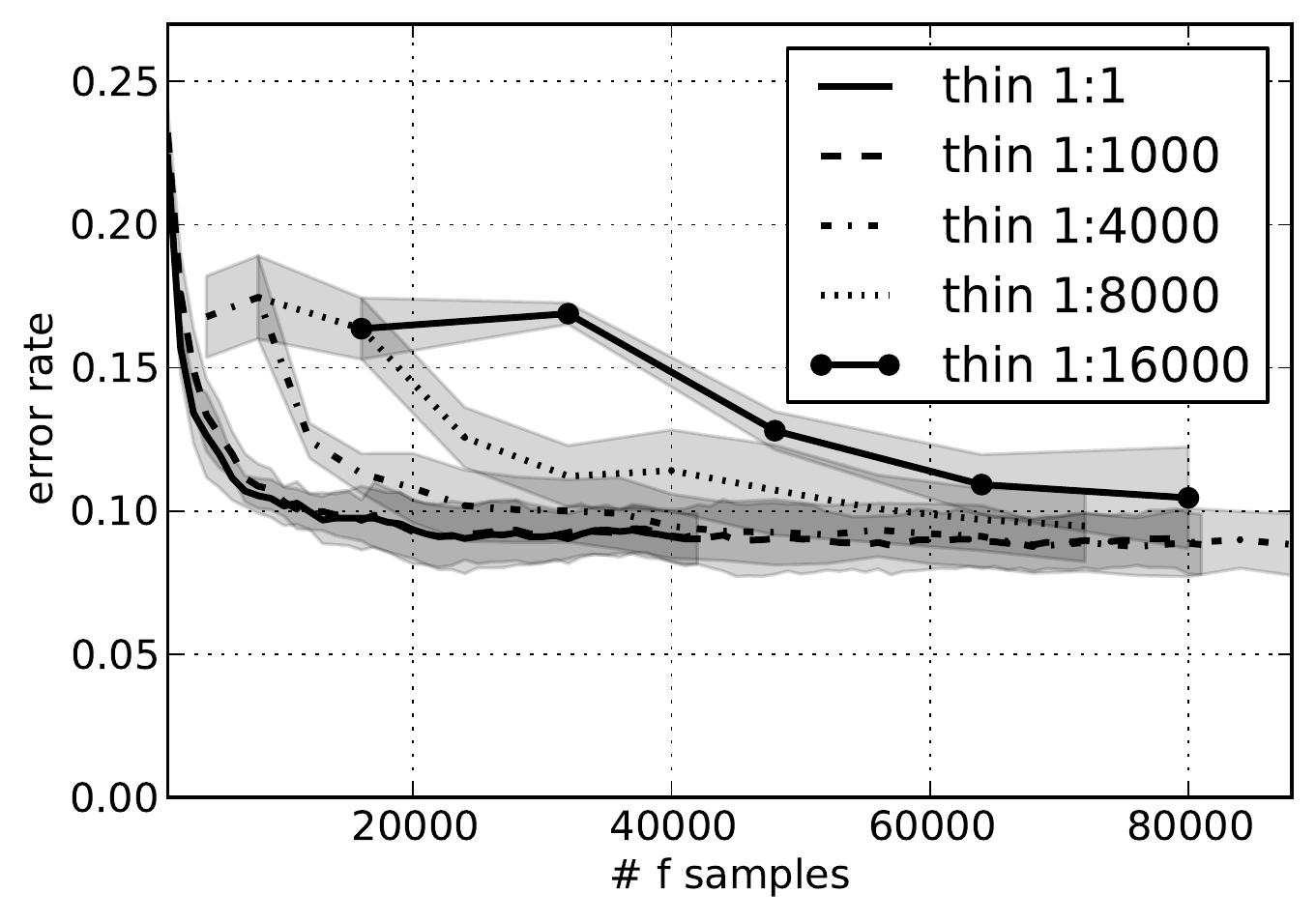}
  \includegraphics[width=0.65\columnwidth]{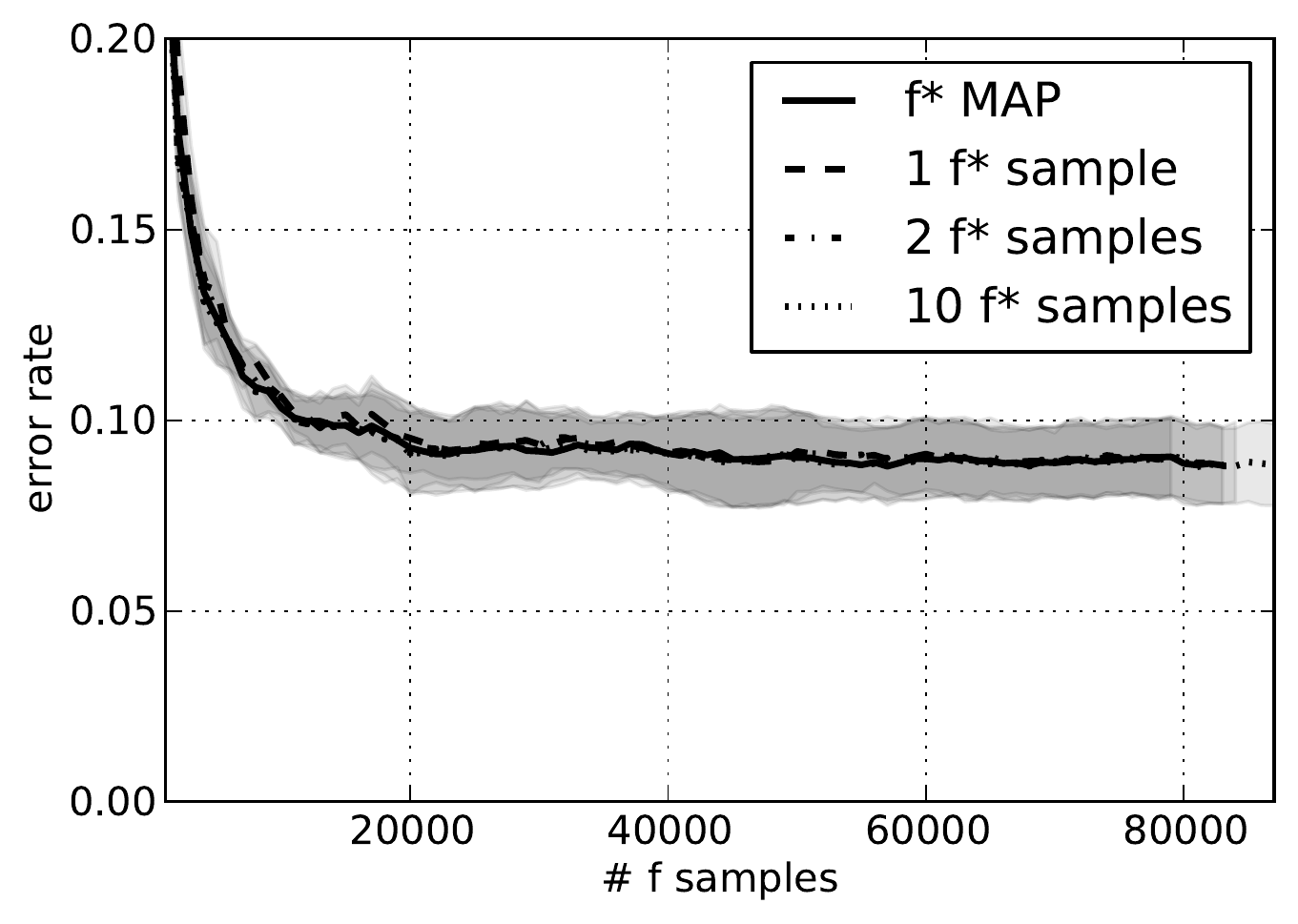}
 \caption{\textbf{top}: Effect of sampling hyperparameters every 1 000 steps versus fixing $h_p=1$, over the full history of $\f$ samples. $\f_*$ MAP scheme, thinning at 1:1 000. \textbf{middle}: Effect of thinning, \ie sampling $\f_* | \f$ more rarely than every $\f$ sample. \texttt{Chunking} task, $\f_*$ MAP scheme, $h_p=1$. \textbf{bottom}: Effect of number of $\f_* | \f$ samples for each $\f$ sample. \texttt{Chunking} task, thinning at 1:1 000, $h_p=1$. \label{fig:mergedpractical}}%
\end{figure*}

\textbf{How many $\f$ samples?} All subplots in figure \ref{fig:mergedpractical} plot the error rate of some configuration against the number of $\f$ samples generated (\ie iterations of the ESS procedure). For all our tasks, the error rate improves until 100 000 iterations, which shows heuristically that sampling histories of at least this length are needed to attain equilibrium for these problems.

\textbf{How many $\f_* | \f$ samples?} $\f_*$ need not be sampled for every $\f$ sample which is generated; to save computing time, we can \emph{thin} and \eg sample $\f_* | \f$ only every 10th $\f$ sample, disregarding the other nine samples entirely. Our exploratory experiments show the following: high thinning rates, such as 1:4 000, seem to have very limited impact on the error rate, cf. figure \ref{fig:mergedpractical} (middle).  Similarly, how many samples $\f_* | \f$ do we need? Do we need any at all, or could we use only the mean of the predictive posterior? This would save computing the predictive variance, which involves a Cholesky matrix inversion, and is called ``$\f_*$ MAP'' here. Figure \ref{fig:mergedpractical} (bottom) answers this: sampling more often does not decrease the error rate. 
These findings are very valuable in practice, and seem to indicate that the predictive posterior is peaked, while the posterior is rather flat, and requires a long MCMC exploration path to be adequately sampled from. Computing time is dictated by the ESS sampling procedure, so performance improvement efforts should clearly aim at obtaining decorrelated posterior samples.

\section{Conclusions and future work}

As a model, GPstruct possesses many desirable properties, discussed in detail above.
Our experiments with sequential data yielded encouraging results: we achieve performance comparable to CRF and exceed SVMstruct in text processing tasks, and exceed CRF in a video processing task.
While GPstruct is theoretically attractive and empirically promising, we have clearly only touched the surface of the model's possibilities. An important limitation preventing the application to larger data sets is the size of the kernel matrix $K$, square in the number of LV. One promising direction is an ensemble learning approach in which weak learners can be trained on subsets of the LV constrained by the underlying MRF (thus with quadratically smaller $K$), and their predictions combined, by Bayesian model combination, into a strong learner. 

\bibliography{bibfile}
\bibliographystyle{unsrt}




  \section*{Acknowledgments}

The authors would like to thank Simon Lacoste-Julien, Viktoriia Sharmanska, and Chao Chen for discussions.

\end{document}